\documentclass[preprint,12pt]{elsarticle}

\usepackage{amssymb}
\usepackage{CJKutf8}
\usepackage{amsmath}
\usepackage{float}
\usepackage{subfigure} 
\usepackage{algorithm}
\usepackage{algorithmic}
\usepackage{booktabs}
\usepackage{graphicx}
\usepackage{times}
\usepackage{latexsym}

\usepackage{bbding}
\usepackage{adjustbox}
\usepackage{amsmath}
\usepackage{graphicx}

\usepackage{longtable}
\usepackage{booktabs}
\usepackage{comment}
\usepackage{multirow}
\usepackage{array}
\usepackage{ulem}
\usepackage{threeparttable}
\usepackage{bbding} 
\usepackage[colorlinks=true, linkcolor=blue, urlcolor=blue, citecolor=red]{hyperref}

\renewcommand{\thefigure}{\arabic{figure}}
\journal{}

\begin{document}
\begin{CJK}{UTF8}{gbsn}
\begin{frontmatter}

\title{LLM-Align: Utilizing Large Language Models for Entity Alignment in Knowledge Graphs}

 \author[a]{Xuan Chen}
 \author[a]{Tong Lu}
 \author[a,b]{Zhichun Wang\corref{cor1}}
 \ead{zcwang@bnu.edu.cn}
 \affiliation[a]{organization={School of Artificial Intelligence, Beijing Normal University},
            city={Beijing},
            country={China}}

 \affiliation[b]{organization={Engineering Research Center of Intelligent Technology and Educational Application, Ministry of Education},
            city={Beijing},
            country={China}}

\cortext[cor1]{Corresponding author}

\renewcommand{\thefigure}{\arabic{figure}}
\begin{abstract}
Entity Alignment (EA) seeks to identify and match corresponding entities across different Knowledge Graphs (KGs), playing a crucial role in knowledge fusion and integration. Embedding-based entity alignment (EA) has recently gained considerable attention, resulting in the emergence of many innovative approaches. Initially, these approaches concentrated on learning entity embeddings based on the structural features of knowledge graphs (KGs) as defined by relation triples. Subsequent methods have integrated entities' names and attributes as supplementary information to improve the embeddings used for EA. However, existing methods lack a deep semantic understanding of entity attributes and relations. 
In this paper, we propose a Large Language Model (LLM) based Entity Alignment method, LLM-Align, which explores the instruction-following and zero-shot capabilities of Large Language Models to infer alignments of entities. LLM-Align uses heuristic methods to select important attributes and relations of entities, and then feeds the selected  triples of entities to an LLM to infer the alignment results. To guarantee the quality of alignment results, we design a multi-round voting mechanism to mitigate the hallucination and positional bias issues that occur with LLMs.
Experiments on three EA datasets, demonstrating that our
approach achieves state-of-the-art performance compared to existing EA methods.
\end{abstract}

\begin{keyword}
Entity Alignment \sep  Large Language Model  \sep Attribute Selection \sep Relation Selection \sep Multi-round Voting Mechanism.
\end{keyword}

\end{frontmatter}

\section{Introduction}
\label{sec:INTRODUCTION}
Knowledge Graphs (KGs) represent structured information of real-world entities, which are widely employed in research fields such as information retrieval \cite{peng2023knowledge,reinanda2020knowledge}, recommendation systems\cite{balloccu2022post,sun2020multi}, image classification \cite{raisi2020selecting,pan2022contrastive}, etc. Most knowledge graphs are developed independently by various organizations, using diverse data sources and languages. As a result, KGs often exhibit heterogeneity, where the same entity may appear in different KGs with varying representations. However, KGs can also complement each other, as information about a single entity may be spread across multiple graphs. To address this heterogeneity and integrate knowledge from different KGs, it's crucial to perform Entity Alignment (EA), which involves matching entities across separate KGs.

The problem of EA has been studied for years, many EA approaches have been proposed. Recently, embedding-based EA has gained considerable attention. Embedding-based EA approaches first learn low-dimensional vector representations of entities, and then match entities in vector spaces. According to the used embedding techniques, embedding-based EA approaches mainly fall into two groups: Translation-based approaches and Graph Neural Network (GNN)-based approaches. Translation-based approaches learn entity embeddings using TransE~\cite{bordes2013translating} and its extensions, such as MTransE\cite{mtranse}, JAPE\cite{jape}, and BootEA\cite{bootea}. GNN-based approaches generate neighborhood-aware entity representations by aggregating the features of their neighbors, representative approaches include GCN-Align\cite{gcnalign}, MuGNN\cite{mugnn}, and AliNet\cite{AliNet}, et al. To further improve the EA results, some approaches explored entities’ attribute information to enhance the entity embeddings, including MultiKE~\cite{multike}, AttrGNN~\cite{AttrGNN} and CEA~\cite{cea}, etc. However, most existing approaches use different techniques to encode information from relational and attributive triples.

Recently, Large Language Models (LLMs) have demonstrated outstanding capabilities in factual question answering\cite{yu2022generate}, arithmetic reasoning\cite{chen2022program}, logical reasoning\cite{pan2023logic}. LLMs hold significant potential for enhancing the understanding of entity relationships and attribute information, facilitating more accurate entity alignment. Several approaches have already proposed to utilize LLMs in EA tasks. AutoAlign\cite{autoalign} employs LLMs to integrate the obtained entity type information as a supervision signal into traditional structure-based methods for joint training. LLMEA~\cite{LLMEA} integrate knowledge from both KGs and LLMs to align entities. ChatEA~\cite{ChatEA} explores the LLMs’ capability for multi-step reasoning to enhance the accuracy of EA. While LLMs have demonstrated their capabilities in EA tasks, there remain challenging issues when applying LLMs to EA:
\begin{itemize}
    \item KGs usually contain large number of both attributive and relational triples of entities, not all of these triples are useful and important for aligning entities. If we take all the triples of related entities as inputs to LLMs, irrelevant triples might disturb the reasoning process of LLMs. 
    \item EA tasks can be formate as judgment questions or multi-choice selection questions for LLMs. The single round decisions made by LLMs sometimes are not reliable, leading to unsatisfying EA results. 
\end{itemize}

To solve the above challenges, we proposed an LLM-based Entity Alignment framework, LLM-Align. LLM-Align first uses any existing EA model as candidate selector, which computes similarities between entities in source and target KGs, and gets candidate alignments by selecting top-k nearest neighbors from the target KG for source entities. Then LLM-Align employs LLMs' reasoning abilities to get the alignment results from the candidates. Specifically, contributions of this work include:
\begin{itemize}
    \item We propose a three-stage EA framework using LLMs. First, alignment candidates are selected using any existing EA model. Next, attribute-based reasoning and relation-based reasoning are performed sequentially, with LLMs utilizing both relational and attributive triples of entities in a consistent manner.
    \item We propose heuristic attribute and relation selection methods for LLM-based EA, which select the most informative attributes and relationships of entities to create concise, effective prompts for LLMs.
    \item We design a multi-round voting mechanism for LLMs to generate reliable EA results. By reordering the input candidates and let LLMs vote for the target entities multiple times, more accurate EA results can be obtained by overcoming the hallucination and positional bias problems in LLMs.
    \item We conduct comprehensive experiments on real-world EA datasets, comparing LLM-Align with several recent approaches. Results demonstrate that LLM-Align effectively enhances the EA performance of both strong and weaker models. When combined with strong models, LLM-Align achieves the best results among all approaches compared.
\end{itemize}

The rest of this paper is organized as follows: 
Section~\ref{sec:RELATED WORK} discusses the related work,
Section~\ref{sec:PRELIMINARIES} formally defines the EA problem,
Section~\ref{sec:PROPOSED METHOD} introduces the details of proposed LLM-Align,
Section~\ref{sec:EXPERIMENTS} presents the experimental results, Section~\ref{sec:CONCLUSIONS} concludes this work.

\section{Related Work}
\label{sec:RELATED WORK}
\subsection{Embedding-based EA}
Embedding-based knowledge graph (KG) alignment methods leverage models like TransE and Graph Neural Networks (GNNs) to learn embeddings for entities, which are then used to identify equivalent entities within vector spaces. Earlier methods primarily focused on utilizing the structural information of KGs for alignment. These include TransE-based approaches such as MTransE, IPTransE~\cite{zhu2017iterative}, and BootEA~\cite{bootea}, as well as GNN-based methods like MuGNN~\cite{mugnn}, NAEA~\cite{naea}, RDGCN~\cite{rdgcn}, and AliNet~\cite{AliNet}.

In these approaches, entity embeddings are learned by incorporating information about entities and their relationships. MTransE encodes the structural details of KGs in separate spaces before transitioning between them. TPTransE and BootEA are iterative alignment methods that expand seed alignments by incorporating newly discovered ones. MuGNN utilizes a multi-channel GNN to learn KG embeddings that are tailored for alignment tasks. NAEA enhances the TransE model by introducing a neighborhood-aware attentional representation for embedding learning. RDGCN employs a relation-aware dual-graph convolutional network, which integrates relation information through attentive interactions between a KG and its dual relation counterpart. AliNet, another GNN-based model, aggregates information from both direct and distant neighborhoods.

To achieve better alignment results, some methods incorporate entity attributes or names from KGs. For instance, JAPE~\cite{jape} employs the Skip-Gram model to perform attribute embedding, capturing attribute correlations within KGs. GCN-Align~\cite{gcnalign} uses Graph Convolutional Networks (GCNs) to encode attribute information into entity embeddings. MultiKE~\cite{multike} adopts a framework that unifies the perspectives of entity names, relations, and attributes to learn embeddings for entity alignment. CEA~\cite{cea} combines structural, semantic, and string-based features of entities, integrating them with dynamically assigned weights.

\subsection{Language Model-based EA}
With the successful application of Pre-trained Language Models (PLMs) across various tasks, some approaches have started leveraging PLMs to model the semantic information of entities in knowledge graph (KG) alignment tasks. For example, AttrGNN~\cite{AttrGNN} employs BERT to encode the attribute features of entities, encoding each attribute and value separately and then using a graph attention network to compute a weighted average of these attributes and values. BERT-INT~\cite{bertint} uses a language model to embed the names, descriptions, attributes, and values of entities, and performs pair-wise neighbor-view and attribute-view interactions to compute the entity matching score. However, these interactions are time-consuming, limiting BERT-INT's scalability to larger KGs. SDEA~\cite{SDEA} fine-tunes BERT to encode the attribute values of an entity into attribute embeddings, which are then processed by a BiGRU to obtain relation embeddings for the entity. TEA~\cite{tea} organizes triples alphabetically by relations and attributes to form sequences and uses a textual entailment framework for entity alignment. TEA inputs entity-pair sequences into a PLM and has the PLM predict the probability of entailment. Like BERT-INT, TEA's pairwise input approach limits its scalability to large KGs. AutoAlign creates attribute character embeddings and predicate-proximity-graph embeddings using large language models.

With the emergence of Large Language Models (LLMs), several approaches have begun exploring their potential for entity alignment (EA). LLMEA~\cite{LLMEA} combines knowledge from KGs and LLMs to predict entity alignments. It first uses RAGAT to learn entity embeddings, which help identify alignment candidates; these candidates are then turned into multiple-choice questions for LLMs to predict the correct alignments. ChatEA~\cite{ChatEA} first utilizes Simple-HHEA~\cite{simple-hhea} to generate alignment candidates and then leverages the reasoning capabilities of LLMs to predict the final results. Both LLMEA and ChatEA exploit the reasoning abilities of LLMs to enhance entity alignment prediction. However, due to the vast number of potential alignments, these methods rely on existing EA techniques to generate alignment candidates, from which LLMs are used to select the final results. The improvements contributed by LLMs in these approaches are, however, somewhat limited.

\section{Problem Definition}
\label{sec:PRELIMINARIES}
\noindent\textbf{Knowledge Graph.} Knowledge Graphs (KGs) represent the structural information of real-world entities through triples in the form $\langle s, p, o\rangle$. There are two kinds of triples in KGs, relational ones and attributive ones. Relational triples capture the relationships between entities, while attributive triples describe the attributes of entities. Formally, we represent a KG as $G=(E,R,A,L,T_{att},T_{rel})$, where $E$, $R$, $A$, and $L$ are sets of entities, relations, attributes, and literals, respectively, $T_{att}\subseteq (E \times A \times L)$ is the set of attributive triples, $T_{rel} \subseteq (E \times R \times E)$ is the set of relational triples.

\noindent\textbf{Entity Alignment.} Given two KGs, a source KG $G = (E,R,A,L,T_{att},T_{rel})$ and target KG
$G^{\prime} = (E^{\prime}, R^{\prime}, A^{\prime},L^{\prime}, T_{att}^{\prime},T_{rel}^{\prime})$, the goal of KG alignment is to identify the equivalent target entity in $E^{\prime}$ for each source entity in $E$.

\section{Proposed Approach}
\label{sec:PROPOSED METHOD}
\begin{figure}[!ht]
    \centering
    \includegraphics[width=1\textwidth]{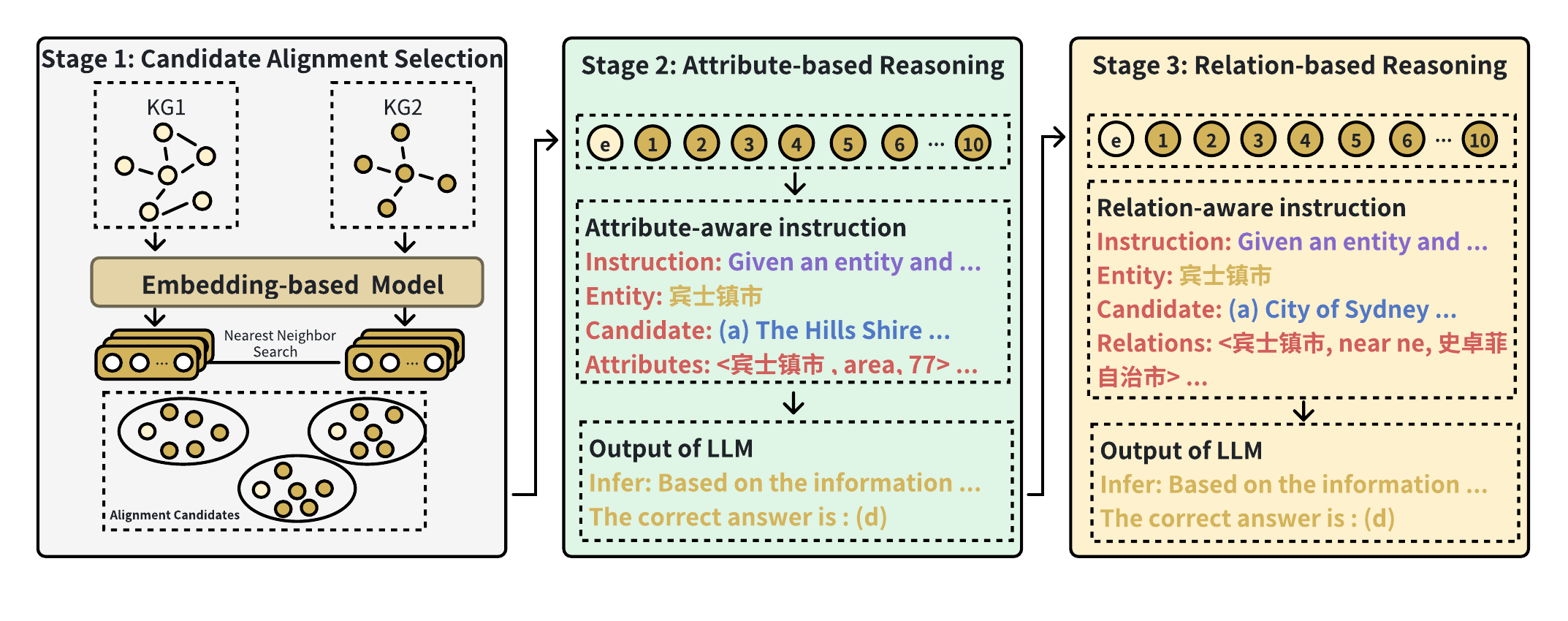}
    \caption{The framework of LLM-Align.}
    \label{gh:llmrerank}
\end{figure}

In this section, we introduce our proposed approach LLM-Align. Figure~\ref{gh:llmrerank} shows the framework of LLM-Align, which works in three stages:
\begin{itemize}
	\item \textbf{Candidate Alignment Selection.} This stage derives entity embeddings by using an existing embedding-based EA model. Subsequently, the nearest neighbor search is performed to obtain candidate alignments for each source entity.
	\item \textbf{Attribute-based Reasoning.} In this stage, LLM-Align identifies and extracts the most informative attributes of entities to construct the inputs for LLMs. A multi-round voting mechanism is employed to conduct multiple parallel reasoning. If a candidate entity is selected in a certain number of reasoning results, it is output as the aligned result; if no entity meets the criterion, LLM-Align moves to the next stage, relation-based reasoning. 
	\item \textbf{Relation-based Reasoning.} This stage selects informative relations of entities to construct the inputs for LLMs. Similar to the previous stage, this stage also uses a heuristic relation selection strategy and employs the multi-round voting mechanism to get the alignment results.	
\end{itemize}

In the following, we introduce our proposed approach in detail. 

\subsection{Candidate Alignment Selection}
\label{sec:candidate alignment selection}

Given a set of source entities $E$ and a set of target entities $E^{\prime}$, candidate alignment selection is to obtain a set of candidate target entities $C_e$ for each source entity $e \in E$, where $C_e\subset E^{\prime}$. More specifically, we employ an existing EA model which generates similarity scores for entity pairs. Target entities having the largest similarities with source entity will be selected as candidates. To control the size of input to LLMs, we let $|C_e|<<|E^{\prime}|$. To avoid introducing any prior biases to LLMs, the previous scores of entities will be ignored in the LLM-based EA reasoning stage. 

\subsection{EA Prompts for LLMs}
\begin{figure}[!ht]
    \centering
    \includegraphics[width=0.95\textwidth]{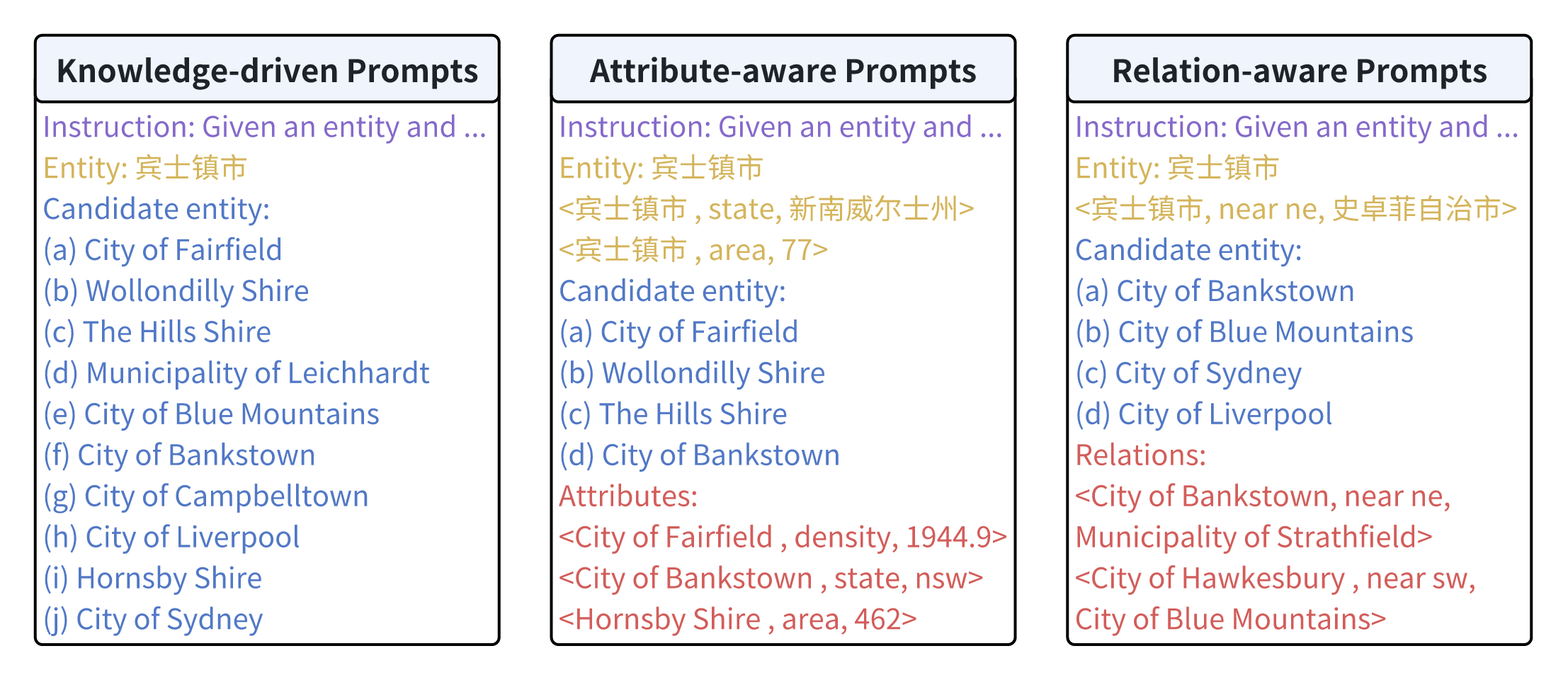}
    \caption{Prompts for LLM-Align.}
    \label{prompts}
\end{figure}

Once a set of candidate alignments are obtained, they will be passed into an LLM to infer the final results. Here we format EA task as single choice selection problems and let LLMs to generate the alignment results based on the given prompts.  
Specifically, the prompt of an EA task contains three parts, including the instruction of EA task, the information of a source entity, and a list of candidate target entities. LLMs are asked to select the most likely target entity for the source entity. Figure~\ref{prompts} shows examples of EA prompts for LLMs. 

Using different background information, we define three types of prompts: (1) Knowledge-driven Prompts: prompts with only entity names; (2) Attribute-aware Prompts: prompts with entity attributes; (3) Relation-aware Prompts: prompts with entity relations.

\textbf{Knowledge-driven Prompts.}
Knowledge-driven prompts only provide LLMs the names of entities, LLMs have to infer the results based on their own knowledge about these entities. For example, as shown in Figure~\ref{prompts}, the knowledge-driven prompt uses entity names as question and options. The correct alignment is (\textit{City of Bankstown},\textit{宾士镇市}), the LLMs are supposed to select the correct target entity \textit{City of Bankstown} from the candidate list.

\textbf{Attribute-aware Prompts.}
\label{sec:Attribute-aware Instructions}
Attribute-aware prompts provide entities' attribute information on the basis of knowledge-driven prompts. It is believed that Entities' attribute information is helpful for LLMs to infer the correct alignment. As shown in Figure~\ref{prompts}, the attribute information of both source entity and candidate target entities are provided with triples. For example, the state and area of \textit{宾士镇市} are given in the prompt, and the population density of \textit{City of Fairfield}, the state of \textit{City of Bankstown}, and the area of \textit{Hornsby Shire} are also provided in the prompt.

\textbf{Relation-aware Prompts.}
\label{sec:Relation-aware Instructions}
Considering attributes of entities might not provide sufficient information to find entity alignments, relation-aware prompts include relation triples of entities to provide evidence for finding alignments. As shown in Figure~\ref{prompts}, the relation triples identifying nearby cities of source and candidate target entities are given in the prompt.

\subsection{Heuristic Attribute and Relation Selection}
In the stages of Attribute-based and Relation-based EA reasoning, LLM-Align identifies and extracts the most informative attributes and relations of entities to construct attribute-aware prompts and relation-aware prompts for LLMs. In this work, we define heuristic rules to guide the selection of attributes and relations. 

\noindent\textbf{Heuristic Attribute Selection. }

Recall that the source KG is $G = (E,R,A,L,T_{att},T_{rel})$ and the target KG is $G^{\prime} = (E^{\prime}, R^{\prime}, A^{\prime},L^{\prime}, T_{att}^{\prime}, T_{rel}^{\prime})$. For a source entity $e\in E$, a set of its candidate target entities $C_e\subset E^{\prime}$ is obtained in the candidate alignment selection stage. To extract informative attributes, we define a metric called \textit{identifiability}, which quantifies the significance of an attribute in distinguishing between different entities. A higher \textit{identifiability} score indicates a more important attribute. Given an attribute $a$, we use $identy_{att}(a)$ to denote the \textit{identifiability} of it. To compute $identy_{att}(a)$, we first compute the \textit{function degree} $fun_{att}(a)$ of $a$:

\begin{equation}
\label{eq:F}
fun_{att}(a)=\frac{|\{h|(h,a,v)\in \{T_{att} \cup T_{att}^{\prime}\}\}|}{|\{(h,v)|(h,a,v)\in \{T_{att} \cup T_{att}^{\prime}\}\}|}
\end{equation}

Then, the frequency $freq_{att}(a)$ of attribute $a$ in the triples of candidate target entities $C_e$ is computed:

\begin{equation}
\label{eq:I}
    freq_{att}(a,C_e)=\frac{|\{h|h\in C_e \land (h,a,v)\in T_{att}^{\prime}|}{|C_e|}
\end{equation}

Finally, the \textit{identifiability} $identy_{att}(a)$ is computed as:

\begin{equation}
\label{eq:D}
    identy_{att}(a, C_e)=fun_{att}(a) \times freq_{att}(a,C_e)
\end{equation}

For each entity in $\{e\}\cup C_e$, we first compute the metrics of \textit{identifiability} for all the attributes of the entity. Then the top-$k$ attributes with the highest \textit{identifiability} are selected, triples of these attributes are added to the attribute-aware prompts for LLMs to decide the true target entity for $e$.

\noindent\textbf{Heuristic Relation Selection.}

Similar to the attribute selection process, we also compute the \textit{identifiability} metrics for relations, and then select the top-$k$ relations to present in relation-aware prompts. For a relation $r$, its function degree $fun_{rel}(r)$ and frequency $freq_{rel}(r,C_e)$ are computed as follows:

\begin{equation}
\label{eq:rfun}
fun_{rel}(r)=\frac{|\{h|(h,r,t)\in \{T_{rel} \cup T_{rel}^{\prime}\}\}|}{|\{(h,t)|(h,r,t)\in \{T_{rel} \cup T_{rel}^{\prime}\}\}|}
\end{equation}

\begin{equation}
\label{eq:rfreq}
    freq_{rel}(r,C_e)=\frac{|\{h|h\in C_e \land (h,r,t)\in T_{rel}^{\prime}|}{|C_e|}
\end{equation}

The \textit{identifiability} $identy_{rel}(r)$ of relation $r$is computed as:

\begin{equation}
\label{eq:idrel}
    identy_{rel}(r, C_e)=fun_{rel}(r) \times freq_{rel}(r,C_e)
\end{equation}

\subsubsection{Multi-round Voting Mechanism}
\label{sec:mrvoting}
When processing long texts, LLMs are affected by positional bias, meaning their performance varies significantly depending on the position of the information within the text. Experiments show that LLMs handle information at the beginning and end of documents more effectively compared to information in the middle. Additionally, hallucinations remain a challenge in practical applications. Factors like model size, training data quality, and the training process contribute to hallucinations. To mitigate the effects of hallucination and positional bias in EA reasoning, we propose a multi-round voting mechanism. 

The multi-round voting mechanism performs multiple independent reasoning. Let the number of votes be $n$, and the size of the candidate entity set be $m$. The method first samples $n$ unique permutations from the $m!$ possible arrangements of the candidate set. The large language model then performs parallel reasoning on the $n$ inputs to generate $n$ independent outputs, and the final answer is determined through a voting mechanism. Let $count(c)$ be the number of times that a target entity $c$ appears in the LLM reasoning results. The target entity $c^{*}$ with the highest $count(c^{*})\geq \lfloor n / 2 \rfloor$ will be chosen as the final alignment result for the source entity. If there is no target entity satisfying $count(c)\geq \lfloor n / 2 \rfloor$, the multi-round voting will not output any alignment results. 

The multi-round voting mechanism described above effectively mitigates the impact of positional bias and hallucinations on large language models, enhancing their accuracy and stability in entity alignment reasoning tasks. Additionally, the multi-round voting mechanism is similar to the concept of ensemble learning, implicitly integrating multiple models, which improves the model's ability to solve complex and difficult problems. In the subsequent experiments, the paper validates the effectiveness of the multi-round voting mechanism in entity alignment tasks, showing that compared to single-round reasoning, the multi-round voting mechanism provides more stable performance in alignment reasoning.

\section{Experiments}
\label{sec:EXPERIMENTS}
\subsection{Experiment Settings}
\subsubsection{Datasets}
In our experiments, we utilize the DBP15K datasets, created by Sun et al.~\cite{jape}. These datasets are derived from DBpedia, a large-scale multilingual knowledge graph that contains abundant inter-language links between different language versions. Specific subsets of the Chinese, English, Japanese, and French versions of DBpedia were selected according to certain criteria. Using these subsets of DBpedia, three cross-lingual EA datasets were built, including Chinese-English (ZH-EN), Japanese-English (JA-EN), and French-English (FR-EN). Details of these datasets are shown in Table.\ref{tb:datasets}.
\begin{table}[h]
    \centering
    \caption{Statistics of DBP15K Datasets}
    \label{tb:datasets}
    \begin{tabular}{lcccccc}
        \toprule
        Dataset & Lang. & Entity & Rel. & Attr. & Rel.triples & Attr.triples \\
        \midrule
        ZH-EN & Chinese & 66,469 & 2,830 & 8,113 & 153,929 & 379,684 \\
        & English & 98,125 & 2,317 & 7,173 & 237,674 & 567,755 \\
        \midrule
        JP-EN & Japanese & 65,744 & 2,043 & 5,882 & 164,373 & 354,619 \\
        & English & 95,680 & 2,096 & 6,066 & 233,319 & 497,230 \\
        \midrule
        FR-EN & French & 66,858 & 1,379 & 4,547 & 192,191 & 528,665 \\
        & English & 105,889 & 2,209 & 6,422 & 278,590 & 576,543 \\
        \bottomrule
    \end{tabular}
\end{table}

\subsubsection{Models Settings}
\noindent\textbf{EA Models for Candidate Alignment Selection.} In the experiments, we use the DERA-R~\cite{dera} and GCN-Align~\cite{gcnalign} models as the base models for candidate entity selection. The DERA-R method and GCN-Align model respectively use textual information and structural information for modeling, both achieving good performance on the Hits@10 metric. Conducting diverse experiments with these two candidate alignment selection models aims to validate the generality of the proposed method, while also allowing this study to further explore how the candidate entity sets retrieved by different information-based modeling approaches impact the final alignment reasoning.

\noindent\textbf{LLMs for EA Reasoning.} The experiment uses Qwen1.5-32B-Chat\cite{bai2023qwen} and Qwen1.5-14B-Chat\cite{bai2023qwen} as reasoning models. Both models can perform efficient reasoning using the vLLM framework on a single 80G GPU. Conducting experiments with models of different scales helps us explore the performance of the method under varying computational resources. At the same time, models of different sizes allow this study to investigate the impact of model scale on reasoning alignment tasks.

\subsubsection{Baselines}
To comprehensively evaluate the performance of this method, we use the following EA models for comparison:
\begin{itemize}
    \item DERA~\cite{dera}: The EA method based on heterogeneous parsing with large language models. This method achieved state-of-the-art (SOTA) performance across multiple datasets.
    \item DERA-R~\cite{dera}: A simplified version of the DERA method, which only uses the heterogeneous parsing module and candidate entity retrieval module, without introducing the related sequential re-ranking module. 
    \item TEA~\cite{tea}: A classic method based on language model modeling, included for comparison to explore the impact and effectiveness of large language models and pre-trained language models on the entity alignment task.
    \item BERT-INT~\cite{bertint}: It is a EA method based on BERT language model, it performs cross-graph interactive modeling of semantic information such as entity names, relationships, and attributes.
    \item HMAN~\cite{hman}: This method uses fine-grained ranking techniques from traditional information retrieval in the final stage to re-rank the candidate entity set. 
    \item AttrGNN~\cite{AttrGNN}: It is a representative method for modeling the topological structure of attribute triples. By comparing with this model, this study explores the impact of using attribute information modeling through graph neural networks and large language models on the entity alignment task.
    \item LLMEA~\cite{LLMEA} integrate knowledge from KGs and LLMs to predict entity alignments.
    \item ChatEA~\cite{ChatEA} first use an existing EA model to generate alignment candidates and then leverages the reasoning capabilities of LLMs to predict the final results.
\end{itemize}

\subsection{Overall Results}

Table~\ref{tb:overallresults} shows the overall results on DBP15K datasets. LLM-Align employs GCN-Align and DERA-R as base models for candidate alignment selection, while Qwen1.5-14B-Chat and Qwen1.5-32B-Chat serve as reasoning models. Unlike other approaches, LLM-Align predicts only a single target entity for each source entity, without computing similarity scores using LLMs. Therefore, only Hits@1 metrics for LLM-Align are shown in Table~\ref{tb:overallresults}. The GCN-Align and DERA-R results were reproduced by us, while results of other methods are sourced from their original papers.

Experimental results demonstrate that LLM-Align is highly effective, achieving state-of-the-art performance. In combination with GCN-Align, LLM-Align significantly boosts results: with Qwen1.5-14B-Chat as the base LLM, it increases Hits@1 by 32.9\%, 34.0\%, and 37.3\% on the ZH-EN, JA-EN, and FR-EN datasets, respectively. When using Qwen1.5-32B-Chat, LLM-Align achieves even greater improvements of 34.9\%, 34.7\%, and 38.0\% in Hits@1 across the same datasets.

When using DERA-R and Qwen1.5-14B-Chat, LLM-Align achieves Hits@1 scores of 97.8\%, 95.7\%, and 99.2\% on the ZH-EN, JA-EN, and FR-EN datasets, respectively. With the larger model, Qwen1.5-32B-Chat, performance improves further, reaching 98.3\% on ZH-EN, 97.6\% on JA-EN, and 99.5\% on FR-EN. Compared to the base model DERA-R, LLM-Align effectively boosts Hits@1 scores with both 14B and 32B models, yielding increases of 2.3\%, 0.7\%, and 0.1\% in Hits@1 with the 14B LLM, and gains of 3.2\%, 2.6\%, and 0.4\% with the 32B LLM. Among all baselines, LLM-Align with the 32B model achieves the highest Hits@1 across all three datasets.

\begin{table}[h]
    \centering
    \caption{Overall Results on DBP15K Datasets}
    \label{tb:overallresults}
    \small
    \resizebox{1\linewidth}{!}{
    \begin{tabular}{lcccccc}
    \toprule
    \textbf{Model} & \multicolumn{2}{c}{\textbf{ZH-EN}} & \multicolumn{2}{c}{\textbf{JA-EN}} & \multicolumn{2}{c}{\textbf{FR-EN}} \\
    \cmidrule(lr){2-3} \cmidrule(lr){4-5} \cmidrule(lr){6-7}
     & \textbf{Hits@1} & \textbf{Hits@10} & \textbf{Hits@1} & \textbf{Hits@10} & \textbf{Hits@1} & \textbf{Hits@10} \\
    \midrule   
    GCN-Align & 0.420 & 0.790 & 0.445 & 0.815 & 0.432 & 0.812 \\
    TEA & 0.941 & 0.983 & 0.941 & 0.979 & 0.987 & 0.996 \\
    BERT-INT & 0.968 & 0.990 & 0.964 & 0.991 & 0.992 & 0.998 \\
    HMAN & 0.871 & 0.987 & 0.935 & 0.994 & 0.973 & 0.998 \\
    AttrGNN & 0.796 & 0.929 & 0.783 & 0.921 & 0.919 & 0.978 \\
    DERA & 0.968 & 0.994 & 0.967 & 0.992 & 0.989 & 0.999 \\
    DERA-R & 0.955 & 0.992 & 0.950 & 0.989 & 0.991 & 1.000 \\
    LLMEA & 0.898 & 0.923 & 0.911 & 0.946 & 0.957 & 0.977\\
    ChatEA & -- & -- & -- & -- & 0.990 & 1.000 \\
    \midrule
    LLM-Align(GCN-Align-Qwen14B) & 0.749 & -- & 0.785 & -- & 0.805& --\\
    LLM-Align(GCN-Align-Qwen32B) & 0.769 & -- & 0.792 & -- & 0.812 & --\\
    LLM-Align(DERA-R-Qwen14B) & 0.978 & -- & 0.957 & -- & 0.992 & -- \\
    LLM-Align(DERA-R-Qwen32B) & \textbf{0.983} & -- & \textbf{0.976} & -- & \textbf{0.995} & -- \\
    \bottomrule
    \end{tabular}
    }
\end{table}

\subsection{Ablation Study}

In this paper, we conducted a series of ablation experiments on base models of different scales (14B and 32B) to gain deeper insights into the effectiveness and significance of each module within the framework. The ablation studies focused on the Attribute-based Reasoning (\textbf{AR}), Relation-based Reasoning (\textbf{RR}), and Multi-round Voting (\textbf{MV}) components. Table~\ref{tab:ablationstudy} shows the results of ablation study.

\textbf{LLM-Align without AR Module.} In ablation experiments with the 14B model, the Hits@1 metric dropped by 16.1\%, 15.3\%, and 14.0\% on the three datasets, respectively. This result demonstrates that, when performing reasoning alignment with a smaller model, the absence of the Attribute-based Reasoning module significantly degrades EA performance, underscoring its importance within the framework. With the 32B model, Hits@1 decreased by 11.1\%, 11.1\%, and 2.2\% across the three datasets. Although the reduction was less severe compared to the 14B model, removing the Attribute-based Reasoning module still had a substantial negative effect on reasoning alignment performance relative to the full framework.

\textbf{LLM-Align without RR Module.} After removing the Relation-based Reasoning module from the complete reasoning framework, experimental results showed a performance drop in both the 14B and 32B models, though the degree of decline varied significantly. In the 14B model, performance decreased by an average of 15.7\% across the three datasets, whereas in the 32B model, performance remained nearly stable, with only a slight decline of 1\%-3\%.

\textbf{LLM-Align without MV Module.} The Multi-round Voting module was designed to address issues such as hallucinations and positional bias that can occur during large language model generation. By shuffling option order and requiring an answer to receive a majority of votes across multiple rounds, it enhances the model’s confidence in aligning the correct entities. After removing this module, performance on the 14B and 32B models dropped by an average of 4.3\% and 1.2\%, respectively, highlighting its effectiveness within the reasoning alignment framework. Analysis of the generation results revealed that large language models often make errors in single-round reasoning decisions. However, with the Multi-round Voting module, the model can still arrive at the correct answer even if a mistake is made in one round of reasoning. This insight offers valuable guidance for designing future entity alignment frameworks based on large language models, especially in balancing model size with accuracy.

\textbf{LLM-Align with Only RR Module.} In this experiment, both the Attribute-based Reasoning and Multi-round Voting modules were removed, leaving only the Relation-based Reasoning module. Under this setup, Hits@1 performance on the 14B and 32B models dropped by an average of 22.7\% and 12.8\%, respectively, compared to the complete reasoning alignment framework. Reintroducing the Attribute-based Reasoning and Multi-round Voting modules led to a marked improvement in reasoning alignment performance, with optimal results achieved when all modules were used together. This experiment demonstrates that the modules complement one another, collectively enhancing the effectiveness of reasoning alignment.

\textbf{LLM-Align with Only AR Module.} In this setting, both the Relation-based Reasoning and Multi-round Voting modules were removed. Results showed that with the 14B model, average Hits@1 performance reached 92.3\%, while with the 32B model, it reached 96.8\%. 

The above experiment results show that the performance of LLM-Align improved as RR and MV modules were incrementally added with AR module. Based on these results, it is inferred that the synergistic effect among modules stems from the complementarity of information and enhanced confidence in the information. Adding the Relation-based Reasoning module to the Attribute-based Reasoning module complements attribute information with structural context, while the Multi-round Voting module strengthens confidence in valuable information. Thus, when all three modules operate together, the framework achieves higher performance than with any one or two modules alone.

\begin{table}[]
    \centering
    \caption{Results of Ablation Study}
    \label{tab:ablationstudy}
    \small
    \begin{tabular}{ccc|cc|cc|cc}
    \toprule
    \multirow{2}{*}{\textbf{AR}} & \multirow{2}{*}{\textbf{RR}} & \multirow{2}{*}{\textbf{MV}} & \multicolumn{2}{c|}{ZH-EN} & \multicolumn{2}{c|}{JA-EN} & \multicolumn{2}{c}{FR-EN} \\
                          &                      &                     & 14B         & 32B         & 14B         & 32B         & 14B         & 32B         \\
    \midrule
    \Checkmark                     &\Checkmark                     & \Checkmark                   & \textbf{0.978}       & \textbf{0.983}       & \textbf{0.957}       & \textbf{0.976}       & \textbf{0.992}       & \textbf{0.995}       \\
    \midrule
    \XSolidBrush                     & \Checkmark                    & \Checkmark                   & 0.817       & 0.872       & 0.804       & 0.865       & 0.852       & 0.973       \\
    \Checkmark                     & \XSolidBrush                    & \Checkmark                   & 0.952       & 0.980       & 0.938       & 0.973       & 0.990       & 0.994       \\
    \Checkmark                     & \Checkmark                    & \XSolidBrush                   & 0.918       & 0.964       & 0.926       & 0.968       & 0.954       & 0.986       \\
    \XSolidBrush                     & \Checkmark                    & \XSolidBrush                   & 0.731       & 0.813       & 0.722       & 0.823       & 0.794       & 0.933       \\
    \Checkmark                     & \XSolidBrush                    & \XSolidBrush                   & 0.909       & 0.971       & 0.914       & 0.952       & 0.947       & 0.981      \\
    \bottomrule
    \end{tabular}
\end{table}

\subsection{Analysis on the Impact of Candidate Alignment Orders}

In this subsection, we examine how the order of candidate entities in prompts impacts the final alignment results. Through experiments with different orders, we analyze whether these variations significantly affect EA outcomes. Specifically, we test the following orders: (1) ordered by similarities, arranged in descending similarity score as determined by the candidate selection model; (2) random order, where candidate entities are randomly shuffled; and (3) reverse order, which is the inversion of the original order.

\begin{figure}[ht]
    \centering
    \subfigure[Knowledge-driven Reasoning.]{
      \includegraphics[width=0.45\textwidth]{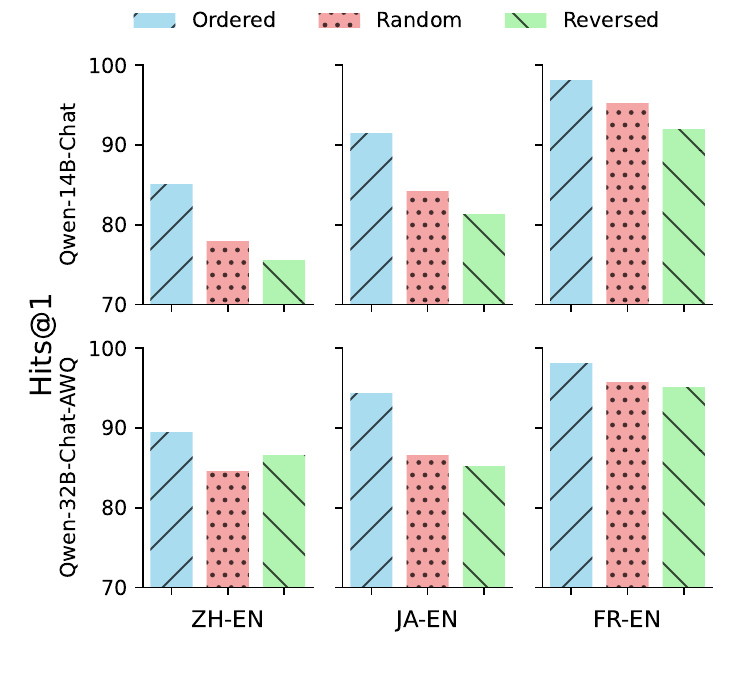}
    }
    \centering
    \subfigure[Attribute-based Reasoning.]{
      \includegraphics[width=0.45\textwidth]{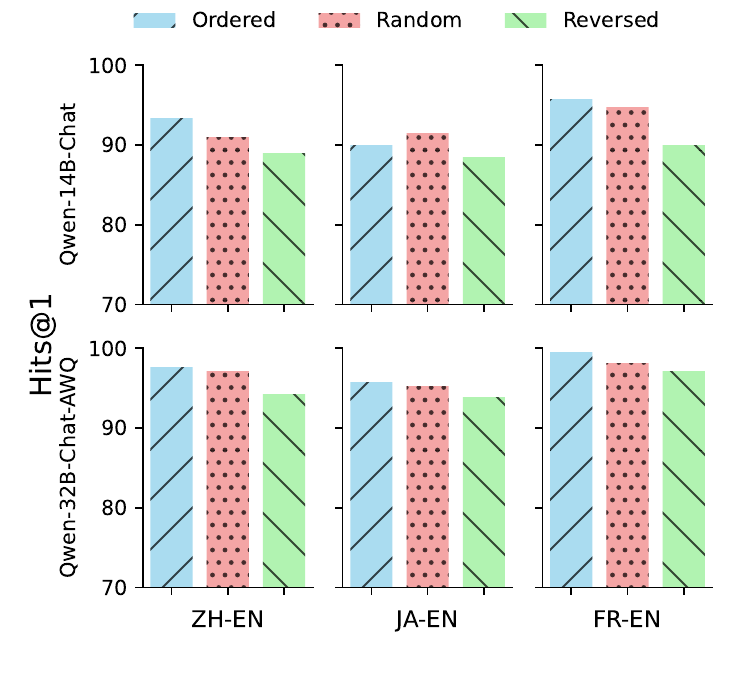}
    }
    \caption{Experimental Results Affected by Positional Bias.}
    \label{fg:position_bias}
  \end{figure}

This experiment was conducted using knowledge-driven prompts and attribute-aware prompts. 
Figure~\ref{fg:position_bias} illustrates the results of different entity orders. The findings indicate that for both the 14B and 32B base reasoning models, retaining the original order from the candidate alignment selection process yielded the best results, followed by random order, while reverse order performed the worst. This suggests that the order information generated during the candidate alignment selection process may contain implicit cues that aid the model's reasoning. The order information reflects the position of the correct answer in the candidate list, highlighting that this position significantly impacts final performance.

When the correct answer is positioned at the front of the list, the likelihood of the model directly identifying it increases due to the order information. For models like DERA-R, which maintain high Hits@1 performance during the candidate matching process, keeping the original order enhances the chances of the correct answer appearing at the top of the candidate list, potentially further boosting the model's performance. Conversely, if the correct entity is located toward the back of the list, this ordering may negatively affect performance by diminishing the model's ability to locate the correct entity. This effect is particularly pronounced for candidate matching models with low Hits@1 performance, as retaining the original order could place the correct answer at the end of the candidate list, thus reducing overall effectiveness.
The experiments above demonstrate that positions significantly influence the final EA results of LLMs, with these models showing a preference for selecting candidates ranked at the top of the list.

\subsection{Analysis on the Impact of LLM Size}
\begin{figure}[ht]
    \centering
    \includegraphics[width=0.7\textwidth]{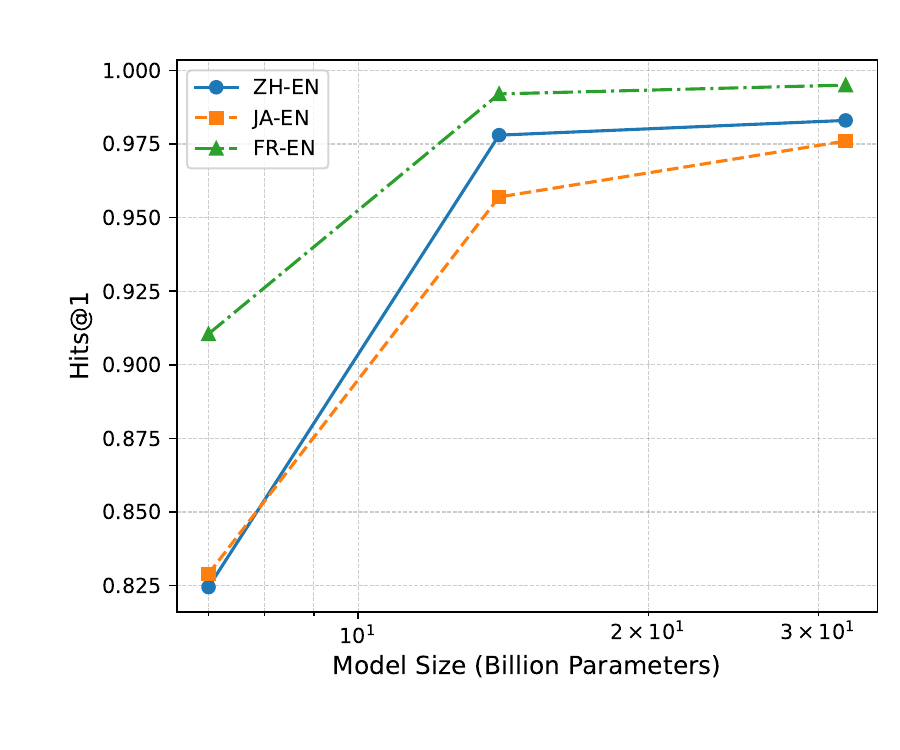}
    \caption{Impact of Model Scale on EA Reasoning.}
    \label{fg:modelsize}
\end{figure}

To investigate the impact of model size on the alignment effectiveness of the proposed method, we conducted experiments using LLMs of varying sizes in the EA inference, specifically 1.5B, 14B, and 32B. The DERA-R model was employed to generate the candidate alignments, with the size of the candidate alignments set to 10. The experimental results are presented in Figure~\ref{fg:modelsize}.

The experimental results indicate that as the model size increases from 1.5B to 32B, performance across all three datasets exhibits a gradual upward trend. This finding reinforces the notion that there is a positive correlation between model size and reasoning alignment performance, suggesting that larger language models can achieve higher Hits@1 scores within the reasoning alignment framework proposed in this study.

Notably, the accuracy of the 1.5B model hovers around 9\% across all datasets, which is close to random selection performance given that the candidate entity set size is 10. Analysis of the model's outputs reveals that the 1.5B model has limited instruction-following capabilities and struggles to accurately understand and execute the reasoning alignment task. This suggests that within the framework proposed in this study, there is a lower limit on model size; below this threshold, the model may be ineffective in performing the reasoning alignment task.

The paper further explores how model size affects the handling of entities with varying difficulty. To precisely differentiate entity difficulty, a grouping method based on similarity calculations during the candidate matching phase was implemented. Specifically, the DERA-R model was first used to match candidate entities for the test set, selecting the top 10 candidates. An entity's difficulty level was determined by whether DERA-R ranked the correct entity first: if the correct entity was not ranked first, it was classified as a high-difficulty entity; otherwise, it was categorized as a low-difficulty entity.

\begin{figure}[ht]
    \centering
    \subfigure[Low-difficulty Entities]{
      \includegraphics[width=0.47\textwidth]{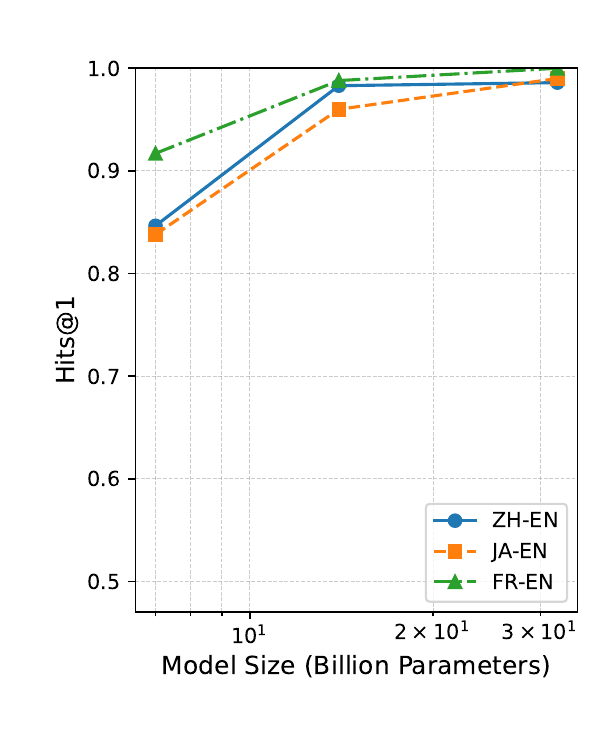}
    }
    \centering
    \subfigure[High-difficulty Entities]{
      \includegraphics[width=0.47\textwidth]{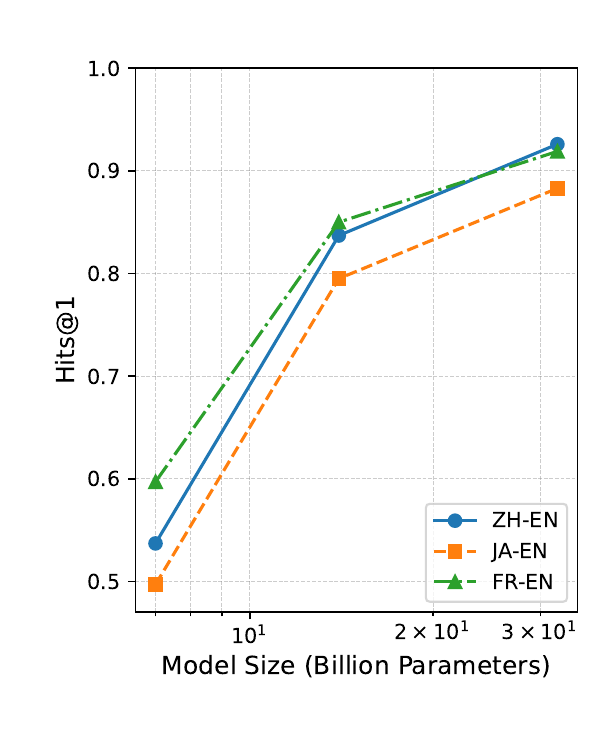}
    }
    \caption{Impact of Model Scale on EA for high-difficulty entity and low-difficulty entity}
    \label{fg:modelscale2}
  \end{figure}

To ensure the stability and reliability of the results, 300 samples were randomly selected from both the high-difficulty entity set and the low-difficulty entity set, and the experiments were repeated three times on these samples. The average result was taken as the final outcome. The experimental results are shown in Figure~\ref{fg:modelscale2}. The results indicate that as the model size increases, the Hits@1 performance improvement for the high-difficulty entity set is greater than for the low-difficulty entity set. These findings align with intuition, as larger models are better equipped to handle more complex and challenging entities.

Although expanding the model size significantly improves its ability to handle high-difficulty entities, we observed that overall performance on the low-difficulty entity set remains higher than on the high-difficulty set. This phenomenon reveals consistency between the reasoning alignment process and the candidate matching process, indicating that even when the candidate matching process has already correctly identified the aligned entity, the reasoning process can maintain or even improve performance.

This consistency suggests that for entities already ranked first in the candidate matching process, the reasoning alignment process can further enhance the accuracy without compromising the original performance. However, for entities not ranked first during the candidate matching process, although the reasoning alignment process faces certain challenges, it often resolves some of the uncertainties left by the candidate matching process, thus substantially improving overall performance.

\subsection{Analysis on the Impact of Candidate Size}

This section aims to explore the ability of the reasoning alignment framework to handle candidate entity sets of varying sizes. By adjusting the size of the candidate entity set, five different scales were used: 10, 20, 30, 40, and 50. The goal was to observe how the accuracy of the proposed framework changes when faced with different numbers of candidate entities. These experiments were conducted on two large language models with sizes of 14B and 32B, using the DBP15K dataset across three datasets, to obtain comprehensive experimental results.

In this experiment, we removed all three modules from the reasoning alignment framework, using only knowledge-driven prompts for LLMs. This setup allows for an accurate evaluation of how the size of the candidate entity set impacts the final results, free from the influence of individual or combined optimizations of the framework's specific modules. To ensure stability and reliability, we extracted 500 samples from the test set, ensuring that the correct entity was included in the top-k candidate entities. The values of top-k were set at 10, 20, 30, 40, and 50, and three sampling experiments were conducted for each candidate set size, with the final result being the average of these three experiments. Natural language instructions used were based on the knowledge-driven instructions proposed in the second section of this chapter.

\begin{figure}[ht]
    \centering
    \subfigure[Qwen-14B-Chat]{
      \includegraphics[width=0.47\textwidth]{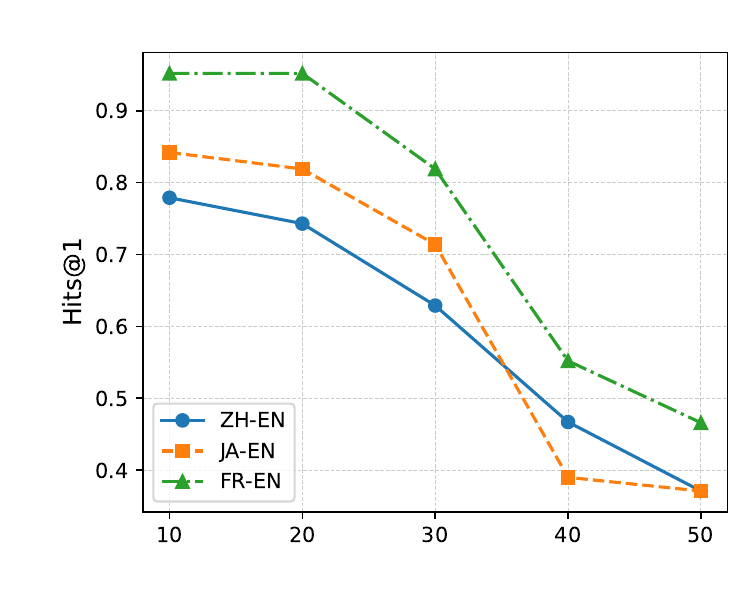}
    }
    \centering
    \subfigure[Qwen-32B-Chat]{
      \includegraphics[width=0.47\textwidth]{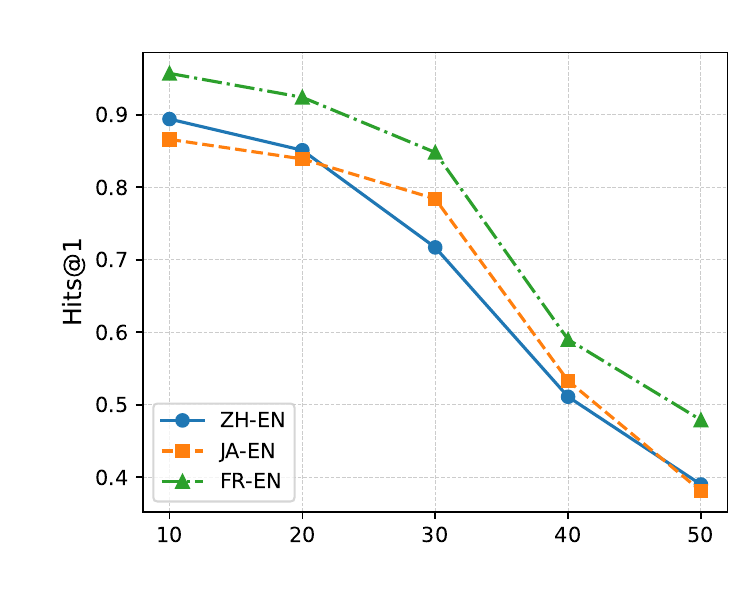}
    }
    \caption{The impact of the number of candidate entities on the Hits@1 metric.}
    \label{fg:candidatesize}
  \end{figure}

The experimental results, shown in Figure~\ref{fg:candidatesize}, were consistent for both the 14B and 32B models. As the number of candidate entities increased, Hits@1 performance showed a downward trend. This trend aligns with expectations: as the number of candidate entities grows, the large language model must make judgments and inferences over more options. This can cause the model's attention to be dispersed across more irrelevant entity options, thereby affecting its ability to correctly identify the target entity.

Overall, the 32B model demonstrated superior performance in this experiment, further validating the importance of model size in improving reasoning alignment performance. However, it is noteworthy that in the FR-EN dataset, when the number of candidate entities was 20, 40, or 50, the performance of the 32B model was on par with, or even slightly lower than, that of the 14B model. Upon analyzing the incorrect samples, we found that most of these errors involved entities with similar names. The 14B model tended to directly output the similar entity as the correct answer, while the 32B model, relying on its own knowledge, attempted to analyze and infer between these similar entities. However, during this inference process, issues arose, leading the model to ultimately arrive at the wrong answer.

The experimental results, as shown in Figure~\ref{fg:candidatesize} indicate that for both 14B and 32B scale models, the Hits@1 performance decreases as the number of candidate entities increases.
This trend is expected, as the large language model has to reason about more options, potentially diverting attention to more unrelated entities, which affects the judgment of the correct entity.
Overall, the 32B scale model performed better in this experiment, further validating the importance of model scale in improving inference alignment performance.
It is noteworthy that on the FR-EN dataset, the performance of the 32B scale model is equal to or slightly lower than that of the 14B scale model when the number of candidate entities is 20, 40, and 50.
The erroneous samples revealed that most of these errors are due to entities with similar names. The 14B model directly output the similar entity as the correct answer, while the 32B model attempted to analyze and infer based on its own knowledge, resulting in incorrect answers.

\section{Conclusions}
\label{sec:CONCLUSIONS}
In this work, we propose an entity alignment approach based on large language models, termed LLM-Align. LLM-Align constructs EA prompts that include candidate alignments for LLMs, leveraging their reasoning capabilities to produce final alignment results. The method employs heuristic techniques to identify key attributes and relationships of entities, incorporating the selected triples into prompts for the LLMs. To ensure high-quality alignment, we design a multi-round voting mechanism that mitigates position bias and addresses hallucination issues associated with LLMs. Experiments conducted on three EA datasets demonstrate that our approach effectively enhances the EA results of existing models. Compared to baselines, LLM-Align achieves the best performance when paired with a robust EA model for candidate selection.

\section*{Acknowledgment}
This work was supported by the National Natural Science Foundation of China (No. 62276026).

\bibliographystyle{elsarticle-num-names} 
\bibliography{LLM-Align}

\end{CJK}
\end{document}